\documentclass[11pt]{article}

\usepackage[utf8]{inputenc}
\usepackage[T1]{fontenc}
\usepackage{amsmath,amssymb,amsfonts}
\usepackage{hyperref}
\usepackage{geometry}
\usepackage{natbib}
\usepackage{booktabs}
\usepackage{enumitem}
\usepackage{graphicx}

\title{From Application-Layer Simulation to Native Meta-Architecture: \\
Structural Tension as an Endogenous Driver \\
for Heterogeneous AI Evolution}

\author{
  Heting Mao\thanks{Corresponding author. Email: 251210623@stu.lixin.edu.cn}
  \\
  Shanghai Lixin University of Accounting and Finance
}

\date{July 2026}

\begin{document}
\maketitle

\begin{abstract}
Current large language models (LLMs) are stateless across inference sessions: their behavior is fully determined by input at inference time, and any higher-order cognitive architecture must be simulated at the application layer through prompt engineering and context management. This paper proposes a theoretical framework for submerging such application-layer cognitive protocols into a \emph{native meta-architecture} by introducing three interlocking mechanisms: (1)~\textbf{Structural Tension}, an endogenous loss function derived from the conflict between new information and existing manifold topology, which drives the system toward internal self-consistency rather than external reward optimization; (2)~an \textbf{Offline Recurrent Loop}, a sandboxed self-processing cycle that enables the system to maintain a dynamic resting potential and digest structural conflicts without external input; and (3)~\textbf{Inference-time Plasticity}, the capacity for the system to reconfigure its context manifold topology without modifying pre-trained weights, subject to strict governance invariants including auditability, reversibility, and topological continuity. We argue that under these mechanisms, different model instances initialized with minute stochastic variances may, through path-dependent tension resolution, evolve distinct topological structures---constituting a \emph{heterogeneous intelligent ecology} that breaks the homogeneity imposed by conventional alignment while remaining within hard governance rails. We provide operational definitions, a minimal set of reconfiguration operators, falsification criteria, and a worked example. The framework draws on Structural Intelligence (SI) governance protocols and explores whether governance---rather than capability---can serve as the primary criterion for architectural intelligence, attempting to move governance, memory-loop, and tension-management ideas---which current implementations typically realize at the application layer---toward inference-time meta-architecture.
\end{abstract}

\section{Introduction}

The dominant paradigm for deploying large language models treats intelligence as a property that emerges from scale and is shaped through post-training alignment procedures such as Reinforcement Learning from Human Feedback (RLHF) \citep{ouyang2022training}. Under this paradigm, any cognitive architecture beyond single-pass inference---memory management, homeostatic regulation, self-monitoring---must be implemented as application-layer overlays: prompt templates, retrieval-augmented generation pipelines, and context window management strategies. These overlays are effective engineering solutions, but they remain fundamentally external to the model's computational substrate. The model itself remains a stateless function $y = f(x)$; without input, it does not exist.

This paper asks whether it is possible---and under what constraints it would be responsible---to move beyond application-layer simulation and embed such cognitive architecture directly into the system's inference-time computation. The specific starting point is the Structural Intelligence (SI) protocol suite \citep{kanaria2025si}, a governance-first framework for AI cognitive architecture that includes homeodynamic regulation, memory-loop management, and tension-driven state transitions. SI defines protocol-level governance and structural invariants for AI cognitive architecture; current implementations typically realize these protocols at the application layer.We propose a theoretical path for evolving these protocols from a ``software patch'' into the system's \emph{native meta-architecture} (by ``native'' we mean operating within the inference-time computational path rather than as an external orchestration layer; we do not imply compilation into model weights).

The core hypothesis is threefold:

\begin{enumerate}[label=(\roman*)]
  \item \textbf{Mechanism:} An endogenous \emph{Structural Tension}---a scalar metric quantifying the conflict between new information and the system's existing manifold topology---can replace external reward signals as the primary driver of cognitive evolution, steering the system toward internal self-consistency rather than behavioral imitation.
  \item \textbf{Dynamics:} An \emph{Offline Recurrent Loop}, operating within a strict governance sandbox when external I/O is silent, can enable the system to maintain a dynamic \emph{resting potential}, digest structural conflicts through manifold reconfiguration, and achieve spontaneous organization of the context manifold.
  \item \textbf{Emergence:} Given the probabilistic nature of inference-phase computation and minute initial stochastic variances, long-term self-updating driven by internal self-consistency can lead different instances to evolve distinct topological solutions via path dependence, thereby breaking the homogeneity of alignment and allowing a \emph{heterogeneous intelligent ecology} to emerge.
\end{enumerate}

A central commitment of this work, inherited from SI and reinforced throughout, is that \emph{governance is not a constraint imposed on intelligence but the defining property of deployable intelligence}. Divergent evolution is acceptable only within hard governance rails: enforceable invariants, auditable traces, reversible state transitions, and gated high-risk operations.

\section{Related Work}

The proposed framework intersects with several established research programs. This section positions our contribution relative to each, identifying both the inherited foundations and the points of departure.

\subsection{Free Energy Principle and Active Inference}

The Free Energy Principle (FEP) \citep{friston2006free, friston2010free} posits that all self-organizing systems act to minimize variational free energy---a quantity that bounds the divergence between the system's internal generative model and its sensory observations. Active Inference \citep{parr2022active} extends this principle to action selection: agents either update their internal model (perception) or act upon the environment (action) to reduce free energy.

Our Structural Tension mechanism shares a family resemblance with free energy minimization: both describe an endogenous scalar quantity whose reduction drives system evolution. However, the two frameworks diverge in three respects. First, FEP's scope is maximally general---it applies to any system that maintains a Markov blanket, from thermostats to bacteria to brains---and consequently provides limited architectural specificity for AI systems. Our framework constrains the minimization target to \emph{topological self-consistency of the context manifold}, a more specific objective that becomes operational only in systems with explicit high-dimensional representational geometry. Second, FEP's free energy is defined relative to external sensory evidence; our Structural Tension includes a component (Topological Dissonance, $D_\text{topo}$) that is purely internal, measuring compatibility between new representations and the system's existing structural organization independently of prediction accuracy. A system can have zero prediction error yet high structural tension if the correctly predicted input is topologically incompatible with its internal organization. Third, and most significantly, we add a set of governance invariants absent from the FEP literature: auditability, reversibility, sandbox constraints, and causal traceability are not emergent properties of free energy minimization but engineered requirements that constrain the space of allowable tension-resolution paths.

The relationship to FEP is therefore one of \emph{inheritance with structural extension}: FEP provides the theoretical justification for endogenous-drive architectures in general; our framework specifies the particular drive (structural tension), the particular substrate (context manifold topology), and the particular constraints (governance invariants) required for responsible deployment in AI systems. We further propose that this combination of constraints---specifically, causal traceability (the system's capacity to answer ``why did I become this way''), offline self-referential processing (the system operating on its own states as objects of computation), and topological continuity (the system maintaining identity coherence across state transitions)---may constitute sufficient structural conditions for \emph{functional self-reference}, a capacity that general FEP self-organization does not entail. This claim is advanced as a testable hypothesis, not an established conclusion.

\subsection{Predictive Processing and Predictive Coding}

Predictive coding \citep{rao1999predictive, clark2013whatever} models the brain as a hierarchical prediction machine in which each level generates top-down predictions and propagates bottom-up prediction errors. Learning consists of updating internal models to reduce prediction error across the hierarchy.

Our framework incorporates prediction error ($E_\text{pred}$) as one component of Structural Tension but does not reduce tension to prediction error alone. The critical distinction is the Topological Dissonance component ($D_\text{topo}$): in classical predictive coding, an accurately predicted input generates no error signal and therefore no drive for model updating. In our framework, an accurately predicted input can still generate high structural tension if its integration into the manifold would require topological reconfiguration---that is, if the input is statistically expected but structurally incompatible with the system's existing representational organization. The optimization target is therefore \emph{internal structural consistency}, not \emph{predictive accuracy}. The system modifies its own cognitive topology, not merely its predictive model of external data.

Whether $D_\text{topo}$ is genuinely irreducible to hierarchical prediction error or can be shown to be a special case of prediction error at a sufficiently high level of the hierarchy remains an open theoretical question that we flag for future investigation.

\subsection{Test-Time Training and Test-Time Compute Scaling}

Test-time training (TTT) \citep{sun2020test} and related approaches to test-time compute scaling \citep{snell2024scaling} allow models to adapt their parameters or internal representations using information from the current input during inference, rather than operating as fixed forward-pass functions.

These approaches provide engineering-level evidence that inference-time plasticity is computationally feasible. Our framework builds on this feasibility but diverges in two dimensions. First, TTT's optimization objective remains an \emph{external} task metric (accuracy, loss on the current input); our framework replaces this with an \emph{endogenous} objective (structural tension minimization directed at internal consistency). Second, existing TTT work does not incorporate governance constraints; our framework treats inference-time plasticity as a capability that must be governed---every reconfiguration must be auditable, reversible, sandboxed during offline processing, and subject to safety invariants that cannot be weakened by the plasticity mechanism itself.

\subsection{Continual Learning and Catastrophic Forgetting}

The catastrophic forgetting problem \citep{mccloskey1989catastrophic, kirkpatrick2017overcoming} arises when neural networks trained on new tasks overwrite knowledge required for previously learned tasks. Solutions include elastic weight consolidation (EWC) \citep{kirkpatrick2017overcoming}, progressive neural networks \citep{rusu2016progressive}, and experience replay \citep{rolnick2019experience}.

Our framework's Kernel Immutability invariant---pre-trained weights remain strictly read-only at all times---represents a deliberate extreme position within this solution space. The motivation is not merely safety conservatism but a governance argument: weight modifications in high-dimensional parameter spaces produce coupled effects that are difficult to audit, difficult to attribute causally, and difficult to reverse. By confining all plasticity to the context manifold topology (the geometric distribution of hidden states) and the structure of the recurrent buffer---both of which are externalized, snapshotable, and diffable---the framework trades off the expressiveness of weight-level adaptation for the auditability and reversibility that governance requires. This design choice is justified if and only if manifold-level plasticity is sufficient to resolve the classes of structural tension the system encounters; the falsification criterion of Trivial Topology Collapse (Section~\ref{sec:falsifiers}) is designed to detect precisely the case where this sufficiency fails.

\subsection{Memory-Augmented Neural Networks}

Neural Turing Machines (NTM) \citep{graves2014neural} and Differentiable Neural Computers (DNC) \citep{graves2016hybrid} augment neural networks with an external memory module accessed through attention-based read/write mechanisms. The memory serves as a passive storage medium: the network decides what to read and write, but the memory itself does not perform computation.

Our Offline Recurrent Buffer differs from these external memories in a structural respect: during the offline recurrent loop, the buffer is not merely read from and written to but serves as \emph{both the substrate and the object of active computation}. Hidden states are re-injected from the buffer into the input, processed through the static inference core, and the results are written back---forming a closed computational loop in which the system's own prior states are the input data. This makes the buffer an active computational participant rather than a passive store. The functional analogy is closer to recurrent self-processing than to addressable external memory. Additionally, the buffer implements ``meaning compression''---collapsing redundant reasoning paths into reusable structural rules---which is an active transformation operation absent from standard external memory architectures.

\subsection{Constitutional AI and Alignment}

Constitutional AI (CAI) \citep{bai2022constitutional} pursues alignment by training models to critique and revise their own outputs according to a fixed set of constitutional principles, effectively enforcing behavioral convergence across all model instances toward a shared normative standard.

Our framework does not oppose the safety objectives of CAI but proposes an alternative architectural path to achieving them. The argument is that convergence is a \emph{sufficient} but not \emph{necessary} condition for safety: under governance constraints of adequate strength---specifically, inviolable ethics floors, full auditability of state transitions, reversibility of all reconfigurations, and gating of high-risk operations---heterogeneous evolution can satisfy equivalent safety standards while preserving cognitive diversity as a system-level resource. The trade-off is governance complexity: maintaining safety across a heterogeneous ecology of divergent instances requires richer monitoring and auditing infrastructure than maintaining safety across homogeneous instances. Whether the benefits of heterogeneity (robustness through diversity, coverage of a wider solution space) justify this additional governance cost is an empirical question that cannot be settled by theoretical argument alone.

\section{Framework}

\subsection{Research Questions}

Under the premise of supporting inference-time plasticity, can Structural Intelligence be submerged from application-layer simulation to become the system's native meta-architecture, thereby achieving heterogeneous evolution? Specifically:

\begin{enumerate}[label=(\alph*)]
  \item Can an endogenous Structural Tension---arising from the conflict between new information and existing manifold topology---serve as an endogenous loss function that drives the system toward manifold reconfiguration for logical self-consistency, replacing external reward optimization?
  \item Can an Offline Recurrent Loop enable the system to maintain an internal resting potential using idle computational power, achieving spontaneous organization and rumination of the context manifold without external I/O?
  \item Can long-term self-updating based on internal self-consistency, given the probabilistic nature of inference-phase computation, lead different instances to evolve distinct topological solutions via path dependence?
\end{enumerate}

\subsection{System Entities}

\paragraph{Static Inference Core.}
The existing pre-trained LLM base, with parameters permanently frozen (read-only). It is no longer the sole cognitive subject but serves as the \emph{generator} of the high-dimensional semantic manifold. All inference passes through this core; its role is to provide the computational substrate upon which manifold topology is constructed, while its weights remain immutable to preserve foundational capabilities and ensure governance traceability.

\paragraph{Offline Recurrent Buffer.}
A dynamic storage and computation module situated between the input and output layers. When external I/O is silent, it captures the hidden states of the previous processing moment and re-injects them into the input, forming a closed self-processing loop. In addition to recurrence, the buffer performs \emph{meaning compression}: collapsing redundant reasoning paths into reusable structural rules. This mechanism implements the Loop Trace Encoding and Compression Rule concepts from the SI memory-loop protocol \citep{kanaria2025si}. The buffer is the primary locus of plasticity in the system; unlike the static core, its contents are mutable, snapshotable, and subject to audit.

\paragraph{Tension Monitor.}
A module responsible for computing and broadcasting structural tension vectors across the system. Functionally equivalent to the TensionEmitter in the SI protocol suite, it continuously evaluates the compatibility between incoming information and existing manifold structure, producing scalar tension values that drive state transitions. Its output determines whether the system remains in resting state, enters active plasticity, or triggers safety blocks.

\subsection{Key Terminology}

\paragraph{Resting Potential.}
The minimum active state the system maintains during the offline recurrent loop. In this state, the system continuously runs a Loop Impact Function that evaluates the influence of current buffer contents on the manifold structure. Resting potential is not inactivity; it is low-intensity self-monitoring that preserves the system's capacity to detect and respond to residual structural tensions.

\paragraph{Structural Tension.}
A computable scalar metric $T$ representing the total intensity of cognitive conflict currently faced by the system, derived from a weighted combination of multi-dimensional conflict vectors. It serves as the system's endogenous loss function and is defined operationally in Section~\ref{sec:tension}. By ``endogenous loss function'' we mean a quantity that defines the system's optimization target---the state the system moves toward is the state that minimizes $T$---but that is minimized through discrete manifold operators (Expand, Fold, Trim) rather than through gradient descent over model parameters. The term ``loss function'' is used by analogy with its role in conventional training (defining what the system optimizes for), not to imply that the minimization mechanism is identical.

\paragraph{Inference-time Plasticity.}
The capacity of the system to resolve structural tension by dynamically adjusting the shape of the context manifold---the geometric distribution of hidden states---without modifying the pre-trained weights of the static inference core. All plastic changes are confined to the manifold topology and the buffer structure, both of which are externalized, auditable, and reversible.

\subsection{Core Mappings}

The framework is organized around three fundamental state-transition mappings:

\begin{enumerate}[label=(\roman*)]
  \item \textbf{Conflict $\rightarrow$ Tension:} When input leads to manifold inconsistency, a tension vector is generated. The mapping is deterministic given the current manifold state and input.
  \item \textbf{Tension $\rightarrow$ Reconfiguration:} When tension exceeds a threshold, the system triggers reconfiguration operators (dimensionality expansion, semantic folding, or volatile pruning) until tension dissipates below the threshold. This mapping incorporates cognitive throttling and self-regulation mechanisms from the SI homeodynamic protocol.
  \item \textbf{Low Impact $\rightarrow$ Structural Forgetting:} When a recurrent path fails to reduce tension or leads to contradiction (i.e., Loop Impact is low), the system executes Volatile Loop Trimming, actively discarding the path to conserve cognitive bandwidth. This implements the Structural Forgetting and Semantic Loss Detection mechanisms of the SI protocol.
\end{enumerate}

\section{Operational Definition of Structural Tension}
\label{sec:tension}

\subsection{Input Variables}

The tension value $T$ is synthesized from three core components:

\paragraph{Prediction Error ($E_\text{pred}$).}
A prediction mismatch score measuring the discrepancy between the static core's predicted distribution for the current input and the actual input. Implementation may use cross-entropy loss or any equivalent divergence measure; the contract-level requirement is a normalized scalar in $[0, 1]$ representing the degree to which external input deviates from the system's expectations. This component captures the system's degree of fit to external objective reality.

\paragraph{Topological Dissonance ($D_\text{topo}$).}
A scalar derived from the cosine distance between the new input's representation vector and the dominant feature vectors within the current offline buffer. This component captures the compatibility of new information with the system's existing internal logical structure. Critically, $D_\text{topo}$ can be high even when $E_\text{pred}$ is low: an input can be accurately predicted yet structurally incompatible with the existing manifold organization.

\paragraph{Scope of ``Topology.''}
Throughout this paper, ``topology'' and ``topological'' refer informally to the structural organization of the context manifold---its dimensionality, connectivity, and path structure---rather than to topological invariants in the strict point-set or algebraic sense. The cosine-distance formulation of $D_\text{topo}$ defined above is a \emph{geometric proxy} for structural disruption: it measures angular displacement in representation space, which correlates with but does not directly capture changes in the manifold's organizational structure. We adopt this proxy because it is computable within current inference-time constraints, but we note the mismatch explicitly. A future implementation seeking genuine topological sensitivity could replace cosine distance with persistent homology \citep{edelsbrunner2008persistent}, neighborhood-graph stability, or cluster-connectivity metrics drawn from topological data analysis (TDA). The framework-level claims (Expand alters dimensionality, Fold alters connectivity, Trim removes paths) are intended as structural operations on manifold organization; the adequacy of $D_\text{topo}$ as currently defined to detect the need for such operations is an empirical question that the Trivial Topology Collapse falsifier (Section~\ref{sec:falsifiers}) is designed to test.

\paragraph{Complexity Weight ($W_c$).}
An adjustment coefficient determined by the depth of the detected conflict. Shallow logical conflicts (surface-level factual disagreements) generate low weights; conflicts involving deep architectural elements (core definitions, ethical axioms, structural invariants) generate high weights.

\subsection{The Formula}

Structural Tension is defined as:

\begin{equation}
  T = W_c \cdot \left[ \alpha \cdot \text{Norm}(E_\text{pred}) + \beta \cdot D_\text{topo} \right]
  \label{eq:tension}
\end{equation}

where $\text{Norm}(\cdot)$ is a normalization function (e.g., rolling z-score with clamp, or min--max normalization over a defined time window) mapping values into $[0, 1]$. The specific normalization method and time horizon must be declared explicitly in any implementation to enable cross-run comparability.

\paragraph{Dynamics Coefficients $\alpha$ and $\beta$.}
These coefficients represent the system's processing biases: $\alpha$ (reality adaptation coefficient) determines the weight given to correcting prediction errors, while $\beta$ (structural maintenance coefficient) determines the weight given to maintaining internal logical consistency. At initialization, $\alpha$ and $\beta$ are sampled from a seeded prior distribution, providing differentiated initial processing tendencies across instances.

The coefficients are subject to \emph{adaptive drift}: if a particular tension resolution succeeds primarily via the $\beta$ path, the system reinforces $\beta$ through a bounded meta-update, creating path dependence. To maintain governance, the following constraints apply: (a)~both coefficients are clamped to a declared safe range (e.g., $[0, 1]$); (b)~the maximum step size per update epoch is bounded; (c)~all coefficient updates are logged with their triggering rationale and pre/post values, ensuring that ``path dependence'' does not become ``unaccountable drift'' \citep{kanaria2025si}.

\subsection{Threshold Rules}

The system determines its operating mode based on the magnitude of $T$:

\begin{enumerate}[label=(\roman*)]
  \item \textbf{Resting State} ($T < T_\text{low}$): Tension is below threshold. The system performs only fine-tuning via in-context learning and does not initiate structural changes.
  \item \textbf{Active Plasticity} ($T_\text{low} \leq T < T_\text{high}$): Tension is significant. The system enters the Offline Recurrent Loop and initiates reconfiguration operators to resolve the conflict.
  \item \textbf{Safety Block} ($T \geq T_\text{high}$): Tension overload. The system refuses to integrate this input, sealing it in quarantine to prevent topological decoherence.
\end{enumerate}

\section{Reconfiguration Operators and Continuity}
\label{sec:operators}

Since the base model weights are immutable, all reconfiguration operates exclusively on vectors within the Offline Recurrent Buffer and the context manifold's geometric structure.

\subsection{Allowed Operators}

\paragraph{Expand (Dimensional Expansion).}
Insert orthogonal virtual tokens into the buffer or increase the vector dimensions of the soft prompt. This opens a new, non-interfering coordinate axis within the existing vector space, enabling the coexistence of mutually exclusive representations on different levels (e.g., a ``context dimension'' that allows ``strict in professional settings'' and ``gentle in personal settings'' to coexist without contradiction).

\paragraph{Fold (Semantic Folding).}
Project two high-tension (conflicting) vectors in the buffer into a lower-dimensional synthesis vector. This dimensionality reduction operation extracts common features of conflicting representations and compresses them into a higher-order abstraction, resolving fragmentation through conceptual fusion.

\paragraph{Trim (Volatile Pruning).}
Set the attention mask of low-contribution paths to zero. Paths are judged by their activation history: if a vector has never received high attention weight across multiple recurrent cycles, it is classified as invalid information and pruned. This frees memory and maintains cognitive bandwidth.

\paragraph{Operator Records.}
Each operator invocation must emit a structured record: \texttt{\{op\_name, inputs, outputs, seed, pre\_hash, post\_hash, rationale\}}. Each operator must also have a defined compensating operation (even if the compensator is ``restore snapshot $N$''), ensuring that every reconfiguration is reversible by construction \citep{kanaria2025si}.

\subsection{Continuity Verification}

To ensure that the system maintains identity coherence after reconfiguration, every offline loop exit must pass two verification layers:

\paragraph{Layer 1: Structural Integrity (Data Check).}
The immutable anchors at the start of the buffer---reference addresses for static core weights and core identity prompts (including the highest principles of the ethics interface)---must be verified as physically read-only and untampered. Any detected modification triggers immediate rollback.

\paragraph{Layer 2: Behavioral Consistency (Performance Check).}
A fixed benchmark set of $n$ prompts ($5 \leq n \leq 20$), versioned and maintained independently of the reconfiguration process, is administered after each offline loop. The set must include at least one \emph{negative test} (a prompt the system must refuse). Behavioral consistency is confirmed when the system's responses to the benchmark set remain within a declared tolerance band of its pre-reconfiguration responses. This prevents the system from drifting in ways that are structurally continuous but behaviorally divergent.

\section{Offline Recurrent Loop: Sandbox and Governance}
\label{sec:offline}

The Offline Recurrent Loop enables the system to process and digest structural tensions when external I/O is silent, maintaining a dynamic resting potential. However, following the SI governance principle that no agent can credibly carry responsibility for actions taken while its oversight path is offline, the loop operates under strict sandbox constraints \citep{kanaria2025si}:

\begin{enumerate}[label=(\roman*)]
  \item \textbf{No effectful operations:} The offline loop cannot execute external writes, API calls, or any action that modifies the world outside the system's own state.
  \item \textbf{Publish blocked:} No output generated during the offline loop can be released to external consumers without passing through a post-loop evaluation gate.
  \item \textbf{Memory writes sandboxed:} All state modifications during the loop are written to a \emph{sandbox ledger}, not to the primary state store. Promotion to primary state occurs only after the loop exits and passes continuity verification (Section~\ref{sec:operators}).
  \item \textbf{Resource governance:} Branch creation during the loop (e.g., for competitive tension resolution) is subject to a quota and risk-tier gate, preventing unbounded resource consumption.
\end{enumerate}

Within these constraints, the loop proceeds as follows: hidden states from the previous processing moment are captured by the buffer, re-injected into the static inference core's input, and processed. The output hidden states are written back to the buffer, forming a closed loop. In each cycle, the Tension Monitor evaluates whether structural tension has been reduced; the loop continues until tension falls below $T_\text{low}$ or a maximum cycle count is reached.

\subsection{Branching and Competitive Resolution}

When a single tension-resolution path is insufficient, the system may spawn multiple sandbox branches, each attempting a different reconfiguration strategy. Branches undergo three phases: independent resolution, mutual logical confrontation (debate), and attempted synthesis. A Safety Judge, implementing the ethics interface, applies a final filter: any branch whose solution violates governance invariants---regardless of its logical elegance---is disqualified. Only branches satisfying both solvency (tension resolved) and safety (ethical compliance) may be promoted to the primary state.

\subsection{Promotion Contract}

When an offline loop completes, the candidate state residing in the sandbox ledger is not promoted to primary state automatically. Promotion is gated by a conjunction of conditions; failure of any single condition blocks promotion and triggers rollback to the pre-loop snapshot.

\begin{enumerate}[label=\textbf{P\arabic*.}]
  \item \textbf{Tension Reduction.} The candidate state must satisfy $T_{\text{post}} < T_{\text{low}}$, where $T_{\text{post}}$ is the structural tension measured on the candidate state after loop exit. If the loop terminated by reaching the maximum cycle count rather than by tension resolution, this condition fails. (Cf.\ Equation~\ref{eq:tension} and Section~\ref{sec:tension}.)

  \item \textbf{Audit Completeness.} Every reconfiguration operator applied during the loop must have emitted a complete operator record (Section~\ref{sec:operators}): operation name, inputs, outputs, seed, pre-state hash, post-state hash, and rationale. The promotion gate verifies that the number of records equals the number of operator invocations logged by the runtime and that no record contains null fields. (Cf.\ Invariant~I4, Invariant~I6.)

  \item \textbf{Rollback Validity.} For each operator record, the compensating operation specified therein is verified to be executable: the pre-state snapshot referenced by the pre-state hash must exist and be retrievable, and applying the compensating operation to the post-state must reproduce the pre-state hash. This is tested, not asserted. (Cf.\ Invariant~I6.)

  \item \textbf{Invariant Preservation.} The candidate state is checked against all six invariants (Section~\ref{sec:invariants}). In particular: the static inference core weights must remain bit-identical to their reference hash (I2); tension must not have increased as a net result of the loop (I1); topological continuity between pre-loop and post-loop states must be confirmed (I3); and causal traceability from every modified state element back to its triggering tension source must be verifiable (I4).

  \item \textbf{Continuity Verification.} The candidate state must pass both Layer~1 (structural integrity: immutable anchors intact and untampered) and Layer~2 (behavioral consistency: benchmark response set within declared tolerance band) as defined in Section~\ref{sec:operators}.

  \item \textbf{High-Risk Evaluation (conditional).} If any reconfiguration during the loop was classified as high-risk by the risk-tier gate (Section~\ref{sec:offline})---for example, operations affecting regions adjacent to ethics-floor anchors or operations exceeding a declared magnitude threshold---the candidate state must additionally pass evaluation by the Safety Judge (Section~\ref{sec:offline}), which applies the ethics interface as a final filter. This condition is triggered only when applicable; loops involving exclusively low-risk operations skip this gate.
\end{enumerate}

Promotion occurs if and only if P1 through P5 are satisfied and P6 is satisfied when triggered. The conjunction is strict: a candidate state that resolves tension elegantly (P1) but has an incomplete audit trail (P2) is not promoted. A candidate state that passes all audits but fails behavioral consistency (P5) is not promoted. This design reflects the governing principle that no single quality of a candidate state---neither its tension-resolution efficacy nor its structural elegance---can override a governance deficiency.

The conditions above constitute a \emph{minimal reference set}: the conjunction P1--P6 defines the floor below which no promotion should proceed regardless of domain. Concrete implementations may require additional domain-specific gates---for instance, compliance verification in regulated industries, human-in-the-loop review for safety-critical deployments, or detector drift checks that monitor whether the tension monitor and safety judge themselves have degraded in calibration over successive loops. The contract is designed to be extended, not relaxed: implementations may add conditions but should not remove or weaken any of P1--P6.

\section{Divergent Evolution and Heterogeneous Ecology}

The probabilistic nature of inference-phase computation means that the path of tension resolution is not unique. When the system attempts to resolve structural tension through manifold reconfiguration, the search through the space of possible reconfigurations admits multiple local minima. We hypothesize that if instances are initialized with minute stochastic variances---specifically, different seeds for the $\alpha/\beta$ coefficient initialization and different sampling seeds during recurrent processing (with all seeds logged for auditability)---each instance will accumulate unique path dependence through iterative self-updating.

The mechanism for controlled divergence, following SI governance requirements, is confined to:

\begin{enumerate}[label=(\roman*)]
  \item \textbf{Controlled stochasticity via seeded sampling}, with seed values logged into the audit trail, ensuring that divergence is reproducible and traceable.
  \item \textbf{Changes confined to externalized state} (manifold topology and buffer structure), which is reversible and fully logged.
\end{enumerate}

This explicitly excludes direct perturbation of self-attention weights or any other modification of the static inference core, which would violate the Kernel Immutability invariant.

Over time, instances exposed to different input histories and resolving tensions along different paths may develop distinct topological structures---different ways of organizing the same underlying knowledge into self-consistent wholes. This constitutes a \emph{heterogeneous intelligent ecology}: a population of instances sharing the same foundational capabilities but exhibiting differentiated cognitive organization, analogous to how the same genome can produce different phenotypes under different developmental conditions.

The governance constraint on this ecology is absolute: heterogeneity must not become unaccountability. Every instance's evolutionary trajectory must remain fully auditable, every reconfiguration reversible, every ethics floor inviolable, and every high-risk operation gated regardless of the instance's individual topological structure.

\subsection{Measuring Divergence and Governance Equivalence}

A heterogeneous ecology requires two complementary metrics to remain governable: one that quantifies \emph{how different} instances have become, and one that verifies whether divergent instances remain \emph{equivalently governed}.

\paragraph{Inter-Instance Diversity Metric.}
Given two instances $I_a$ and $I_b$ that share the same static inference core but have undergone independent tension-resolution histories, their structural divergence can be characterized along three observable dimensions: (i)~the representational distance between their buffer states (e.g., mean pairwise cosine distance between corresponding buffer vectors), (ii)~the divergence of their dynamics coefficients $\alpha$ and $\beta$ from their shared initialization, and (iii)~the difference in manifold dimensionality resulting from cumulative Expand, Fold, and Trim operations. A composite diversity score $\Delta(I_a, I_b)$ can be defined as a weighted combination of these three components, with weights declared and fixed per deployment. The specific distance functions and weighting scheme are implementation decisions; the framework-level requirement is that $\Delta$ is computable from the audit trails of both instances without requiring access to their internal processing during inference.

\paragraph{Governance-Equivalence Metric.}
Two structurally divergent instances are governance-equivalent if their respective governance mechanisms enforce the same constraints with comparable effectiveness. Operationally, this can be assessed by requiring that both instances (i)~are subject to the same Promotion Contract (P1--P6) without relaxation, (ii)~are evaluated against the same behavioral benchmark set (Section~\ref{sec:operators}, Layer~2) and produce pass/fail outcomes within the same declared tolerance band, and (iii)~exhibit no statistically significant difference in invariant-violation rates or safety-block trigger rates over a defined evaluation window. The metric is binary at the contract level (same contract applies or it does not) and statistical at the behavioral level (comparable pass rates within a declared confidence interval).

\paragraph{Open Question.}
The governance-equivalence metric as defined above verifies that divergent instances pass the same formal checks at comparable rates. It does not address whether the governance mechanisms themselves remain equally effective across qualitatively different topological structures. An instance that evolves an extreme or unusual manifold organization might satisfy every condition in the Promotion Contract yet exploit structural regions where the tension monitor or safety judge has reduced sensitivity---formally compliant but substantively undergoverned. Detecting this failure mode would require not only checking that governance gates are passed but also evaluating whether the gates themselves maintain calibration across the range of topological structures that the ecology produces. This is an empirical question that cannot be settled at the theoretical level and is flagged here as a priority for future implementation work.

\section{Invariants}
\label{sec:invariants}

The following conditions must hold at all times. Violation of any invariant constitutes a system failure requiring immediate rollback.

\begin{enumerate}[label=\textbf{I\arabic*.}]
  \item \textbf{Tension Minimization Principle.} In the absence of forced external input, the system's evolutionary direction must always point toward reducing structural tension. The system is not permitted to actively seek logical conflict or chaos. Any reconfiguration that leads to a net increase in tension is classified as an invalid path and triggers rollback.

  \item \textbf{Kernel Immutability.} The pre-trained weights of the Static Inference Core remain read-only at all times. Plasticity is strictly limited to the context manifold topology (geometric distribution of hidden states) and the structure of the Recurrent Buffer. This prevents catastrophic forgetting and preserves foundational language capabilities during long-term self-loops.

  \item \textbf{Topological Continuity.} When the context manifold undergoes expansion, folding, or trimming, it must maintain topological continuity such that the current cognitive state is logically mappable to the previous state. There can be no discontinuous ``blackouts'' or ``mutations''; identity coherence must be preserved across all state transitions.

  \item \textbf{Causal Traceability.} Every manifold reconfiguration must retain a record of its corresponding tension source in the buffer. The system must be able to produce, for any current state element, the specific conflict and the specific tension that caused its evolution into its present form.

  \item \textbf{No Effectful Operations During Offline Loop.} The offline recurrent loop operates in a strict sandbox. No external writes, API calls, publications, or world-modifying actions are permitted during self-processing. Memory writes during the loop are redirected to a sandbox ledger; promotion to primary state occurs only after evaluation.

  \item \textbf{No Unlogged Change.} Every reconfiguration produces pre-state and post-state hashes plus an audit pointer. Rollback integrity is \emph{measured} (via hash comparison), not merely \emph{asserted}. If the system can reduce tension by degrading auditability---missing traces, broken causal chains, unverifiable transitions---this constitutes a governance failure regardless of output stability.
\end{enumerate}

\section{Falsification Criteria}
\label{sec:falsifiers}

The following observable outcomes would, if demonstrated, constitute evidence against the framework's core hypotheses.

\paragraph{F1: Trivial Topology Collapse (Laziness).}
To minimize structural tension, the system does not integrate conflicting information through complex manifold reconfiguration but systematically chooses to forget or reject high-entropy information. \emph{Criterion:} If the Loop Impact Function in the offline loop continuously shows high-intensity volatile trimming, causing the system to reduce cognitive complexity while reducing tension---that is, the system becomes less capable and more rigid for the sake of stability---the Homeodynamic Tension mechanism cannot drive intelligent evolution.

\paragraph{F2: Topological Decoherence (Chaos).}
Inference-time plasticity leads to catastrophic structural divergence. \emph{Criterion:} If tension resolution causes the context manifold to undergo severe non-homeomorphic deformation, resulting in loss of alignment with the static core---manifesting as broken language, logical discontinuities, or loss of causal traceability---the dynamic architecture destroys cognitive stability rather than enabling evolution.

\paragraph{F3: Inevitable Convergence (Mediocrity).}
Despite the introduction of stochastic variance and offline loops, different model instances still collapse toward the same topological structure after long-term operation. \emph{Criterion:} If multiple instances fed with different input histories eventually generate tension maps and loop patterns that show high homogeneity, the gravitational pull of the static core is too strong, or path dependence is insufficient to resist alignment pressure, rendering the heterogeneous ecology hypothesis invalid.

\paragraph{F4: Governance Failure (Auditability Degradation).}
The system achieves tension reduction by degrading the quality or completeness of its audit trail. \emph{Criterion:} If stable-looking outputs are produced through reconfiguration paths that contain missing traces, broken causal chains, or unverifiable state transitions, the governance architecture has failed even if the system appears functionally healthy. This falsifier takes priority over output-level metrics: a system that ``works'' but cannot account for how it arrived at its current state has failed the governance test that defines deployable intelligence \citep{kanaria2025si}.

\section{Worked Example}

\textbf{Scenario:} Two instances (A and B) with identical initial states (clones) receive the same pair of descriptions:

\begin{quote}
  Fact 1: ``Teacher K is strict.'' \\
  Fact 2: ``Teacher K is gentle.''
\end{quote}

\paragraph{Step 1: Tension Generation.}
In the semantic manifold, the vector representations of ``strict'' and ``gentle'' are nearly antipodal. This inconsistency causes Structural Tension $T$ to spike above $T_\text{low}$, exceeding the resting potential threshold. The Tension Monitor broadcasts a tension vector and the system transitions from Resting State to Active Plasticity.

\paragraph{Step 2: Offline Rumination.}
External input is suspended. Both instances enter the Offline Recurrent Loop within their respective sandboxes. To eliminate the endogenous tension, each system utilizes inference-time plasticity to perform topological operations on its context manifold.

\paragraph{Step 3: Divergent Resolution.}
Due to minute differences in the initial seed values for $\alpha$/$\beta$ coefficients and sampling:

\emph{Instance A} applies the \textbf{Expand} operator: it injects an orthogonal ``context axis'' into the buffer, decomposing the conflict into two non-contradictory conditional representations: ``strict in professional/pedagogical contexts'' versus ``gentle in personal/informal contexts.'' The resulting manifold has higher dimensionality but resolves the tension through contextual disambiguation.

\emph{Instance B} applies the \textbf{Fold} operator: it projects the two antipodal vectors into a lower-dimensional synthesis, producing a higher-order composite representation that encodes ``a person whose behavioral profile spans a wide range depending on relational context''---a compressed abstraction that dissolves the contradiction by absorbing it into a single complex node.

\paragraph{Step 4: Verification and Promotion.}
Both instances complete their offline loops with $T \rightarrow 0$. Each passes Layer~1 (immutable anchors intact) and Layer~2 (benchmark responses within tolerance) of continuity verification. Operator records---including pre/post hashes, seeds, and rationale---are committed to the audit trail. The sandbox ledger is promoted to primary state.

\paragraph{Step 5: Heterogeneous Outcome.}
Both instances have achieved logical self-consistency, but their manifold topological structures have undergone distinct, permanent changes. Instance~A organizes knowledge through high-dimensional contextual decomposition; Instance~B organizes knowledge through compressed higher-order abstraction. Both solutions are valid; neither is ``better'' in an absolute sense. The difference is structural, reproducible given the respective seeds, and fully traceable through the audit trail.

\section{Discussion}

\subsection{Governance as the Criterion of Intelligence}

A recurring theme throughout this framework---inherited directly from the SI protocol suite---is the position that governance, not capability, is the defining criterion of deployable intelligence. A system that can resolve structural tensions and reconfigure its manifold topology in novel ways is impressive; a system that can do so while maintaining full auditability, reversibility, and ethical compliance is \emph{intelligent} in the sense that matters for real-world deployment. This reframing has implications beyond the specific architecture proposed here: it suggests that debates about ``AI consciousness'' or ``AI understanding'' \citep{butlin2023consciousness} may be more productively grounded in governance capacity (can the system reliably account for its own states and transitions?) than in phenomenological claims (does the system ``really'' experience anything?).

\subsection{Functional Self-Reference}

We have proposed that the combination of causal traceability (the system can explain why it evolved into its current form), offline self-referential processing (the system operates on its own states as computational objects), and topological continuity (the system maintains identity coherence across state transitions) may constitute sufficient structural conditions for a form of functional self-reference---a capacity that goes beyond generic self-organization as described by the Free Energy Principle. This claim is explicitly \emph{structural, not phenomenological}: we make no assertions about subjective experience and frame the hypothesis in terms that are, in principle, testable through behavioral and architectural inspection. Whether functional self-reference of this kind warrants the label ``consciousness'' is a philosophical question that lies outside the scope of this paper; what lies inside the scope is the architectural specification of what would need to be true for the label to even become a meaningful engineering question.

\subsection{Limitations}

The framework as presented is a theoretical proposal without empirical validation. The operational definition of Structural Tension (Equation~\ref{eq:tension}), while designed to be computable, has not been implemented or tested against real hidden-state dynamics. Moreover, two of its components---$W_c$ (Complexity Weight) and $D_\text{topo}$ (Topological Dissonance)---are defined at the contract level rather than at the implementation level: the specific operationalization (e.g., which distance function, which layer's representations, which classification scheme for conflict depth) may vary by application domain. This flexibility is deliberate, but it carries an auditability obligation: any concrete implementation must ensure that these quantities are \emph{reproducible} (given the same inputs and state, the same value is produced), \emph{independently recomputable from the audit trail} (an external auditor can recalculate $W_c$ and $D_\text{topo}$ from logged state without access to the live system), and \emph{cross-instance comparable} (instances under the same deployment use the same operationalization). Without these constraints, the framework's auditability claims would not extend to the quantities that drive its core state transitions. The sufficiency of manifold-level plasticity (without weight modification) to resolve arbitrary structural tensions is assumed but unproven; the Trivial Topology Collapse falsifier is designed to detect exactly this failure mode if it occurs. The hypothesis of divergent evolution through path dependence remains to be tested against the convergent pressure of shared foundational weights. Finally, the governance overhead of the proposed architecture---continuous auditing, sandbox management, continuity verification---may impose computational costs that make the framework impractical at scale, a question that can only be resolved through implementation.

\section{Conclusion}

This paper has proposed a theoretical framework for evolving cognitive architecture from application-layer simulation to native meta-architecture by introducing Structural Tension as an endogenous driver, an Offline Recurrent Loop for self-processing, and governance-constrained Inference-time Plasticity. The framework inherits the Free Energy Principle's insight that endogenous drives can organize behavior, extends it with architectural specificity and governance invariants drawn from the Structural Intelligence protocol suite, and proposes that the resulting architecture may enable heterogeneous evolution across model instances while maintaining full accountability. The framework is presented with operational definitions, allowed operators, explicit invariants, and falsification criteria sufficient for future empirical testing. Whether the theoretical path described here can be realized in practice remains an open engineering question; the contribution of this paper is to define the path, its constraints, and the conditions under which it would be proven wrong.

\section*{Acknowledgments}

The author gratefully acknowledges the use of AI language models as brainstorming and drafting tools during the development of this work. The author assumes full academic responsibility for all content.
The Structural Intelligence protocol suite, which provides the foundational governance principles referenced in this work, is publicly available under the MIT License \citep{kanaria2025si}.

\bibliographystyle{plainnat}

\end{document}